# ViRanker: A BGE-M3 & Blockwise Parallel Transformer Cross-Encoder for Vietnamese Reranking

Phuong-Nam Dang[1, 2], Kieu-Linh Nguyen[2] and Thanh-Hieu Pham[2]

[1] Bank for Investment and Development of Vietnam (BIDV), Hanoi, Vietnam
[2] Posts and Telecommunications Institute of Technology, Hanoi, Vietnam
phuongnamdpn2k2@gmail.com, linhnk@ptit.edu.vn and hieupt@ptit.edu.vn

**Abstract.** This paper presents ViRanker, a cross-encoder reranking model tailored to the Vietnamese language. Built on the BGE-M3 encoder and enhanced with the Blockwise Parallel Transformer, ViRanker addresses the lack of competitive rerankers for Vietnamese, a low-resource language with complex syntax and diacritics. The model was trained on an 8 GB curated corpus and fine-tuned with hybrid hard-negative sampling to strengthen robustness. Evaluated on the MMARCO-VI benchmark, ViRanker achieves strong early-rank accuracy, surpassing multilingual baselines and competing closely with PhoRanker. By releasing the model openly on Hugging Face, we aim to support reproducibility and encourage wider adoption in real-world retrieval systems. Beyond Vietnamese, this study illustrates how careful architectural adaptation and data curation can advance reranking in other underrepresented languages.

## 1   Introduction

Recent advances in retrieval-augmented generation (RAG) [1] and search-based applications have highlighted the importance of robust reranking models in information retrieval pipelines. While dense retrievers can retrieve a broad set of candidate documents, it is the reranker—typically a cross-encoder—that determines the final ordering of results. High-quality reranking is particularly important because users often examine only the top few results [1].

Despite rapid progress in English and other high-resource languages, reranking models for Vietnamese remain underdeveloped. Vietnamese poses unique challenges: its flexible word order, rich use of diacritics, and high degree of syntactic ambiguity often limit the effectiveness of multilingual models. While large pretrained multilingual encoders have made Vietnamese somewhat accessible, they still lag behind dedicated monolingual approaches in accuracy and robustness. This gap has motivated research into Vietnamese-specific models such as PhoBERT [2] and PhoRanker [3], which demonstrate the value of native-language pretraining and task-specific design. However, PhoRanker and similar systems face scalability issues and do not fully exploit recent architectural innovations in efficient long-context modeling.

To address these challenges, we introduce ViRanker (*https://huggingface.co/namdp-ptit/ViRanker*), a reranking model explicitly optimized for Vietnamese information retrieval. Since its public release on Hugging Face, ViRanker has sustained an active user base of approximately 5,000 monthly users, peaking at over 23,000, and consistently outperforms existing reranking approaches on the MMARCO-VI benchmark [4]. ViRanker is a Vietnamese-specific cross-encoder reranker that builds upon the BGE-M3 encoder [6] and incorporates two key modifications: (i) Rotary Position Encoding (RoPE) [7], which better captures word order and positional nuances critical in Vietnamese, and (ii) the Blockwise Parallel Transformer (BPT) [8], which improves computational efficiency and enables effective handling of long documents. In addition, ViRanker leverages a hybrid hard-negative mining strategy that combines BM25 retrieval [9], dense similarity reranking, and Maximal Marginal Relevance (MMR) filtering [10] to ensure robust training.

This paper makes the following contributions:

**A Vietnamese-specific reranker**. We present ViRanker, the first cross-encoder reranker that integrates the BGE-M3 backbone with RoPE and BPT, offering a design tailored to the linguistic and computational challenges of Vietnamese retrieval.

**A refined training pipeline**. We construct a curated 8 GB Vietnamese corpus and develop a hybrid hard-negative mining strategy that yields a large, high-quality training set of 3.5 million triplets.

**Comprehensive evaluation**. We benchmark ViRanker on MMARCO-VI [4], demonstrating strong early-rank accuracy (NDCG@3 = 0.6815, MRR@3 = 0.6641) and competitive deep-rank performance compared to PhoRanker, along with error analysis to highlight model limitations.

**Efficiency and open release**. ViRanker delivers competitive inference efficiency across both high-end and mid-tier GPUs, supporting practical deployment. The model is released openly on Hugging Face to encourage reproducibility and community use.

The rest of the paper is organized as follows: Section 2 reviews related work on reranking and Vietnamese-specific models; Section 3 describes ViRanker's architecture, data preparation, and training procedure; Section 4 presents experimental results and analysis; and Section 5 concludes with key findings and future directions.

## 2  Related Work

Information retrieval (IR) systems typically operate in two stages: an initial retrieval phase that surfaces candidate documents and a reranking phase that reorders them based on deeper relevance modeling. Classical methods such as BM25 [9] remain strong baselines due to their simplicity and efficiency, but they struggle with semantic similarity, particularly in morphologically rich or low-resource languages like Vietnamese.

The advent of pretrained language models transformed reranking. Built upon the Transformer architecture [14], cross-encoder designs such as MonoBERT [11] demonstrated that fine-tuned transformers could substantially improve retrieval accuracy. Multilingual extensions like mBERT [12] and XLM-R [13] further expanded these gains across languages. Refinements such as relative position encoding [15] improved the ability of self-attention to capture structural dependencies, influencing the design of subsequent rerankers. More recent models, including BGE-Reranker-V2-M3 [17] and Gemma-Reranker [16], enhanced scalability and multilingual generalization by refining both backbone encoders and training pipelines.

Beyond architecture, negative sampling strategies have become central to reranker training. Instead of relying on random negatives, hard-negative mining (e.g., BM25 retrieval [9], dense similarity filtering, or Maximal Marginal Relevance [10]) forces models to distinguish between relevant and distractor passages. Techniques such as the Memory Bank [11] extend this idea by maintaining a dictionary of negative embeddings, enlarging the effective training set without requiring large in-batch negatives. An example PyTorch implementation of this mechanism is shown in Fig. 1, where embeddings are queued, updated, and then reused in contrastive loss computation. Building on this idea, Momentum Contrast (MoCo) [19] introduces a dynamic queue updated via momentum rules, which alleviates staleness and ensures that negative samples remain consistent over training.

```python
queue = None

anchor = model(input_ids, attention_mask)
positive = model(input_ids_positive, attention_mask_positive)

if queue is None:
    queue = positive.detach()
else:
    queue = torch.cat([positive.detach(), queue], dim=0)

queue = queue[:max_queue_length]

scores = cosine_similarity(anchor, queue)
labels = torch.arange(scores.size(0), device=anchor.device, dtype=torch.long)

loss = torch.nn.CrossEntropyLoss()

return loss(scores, labels)
```

**Fig. 1.** Illustration of a PyTorch implementation of the Memory Bank mechanism. Embeddings are queued, updated, and then reused as negatives in contrastive loss computation, enabling resource-efficient training in memory-constrained environments.

To scale further, researchers have also introduced new methods for handling long contexts in Transformer models. Rotary Position Encoding (RoPE) [7] integrates both absolute and relative positional information by rotating feature pairs in embedding space, and is lightweight enough to be combined with linear or memory-efficient attention mechanisms [5]. Blockwise Parallel Transformer (BPT) [8] goes further by chunking attention and interleaving it with feed-forward layers, achieving up to 32× longer context lengths compared to vanilla attention and 2–4× more than FlashAttention [5]. The benefits of BPT over other attention methods are illustrated in Fig. 2, which shows its ability to extend the maximum context length significantly across different hardware setups.

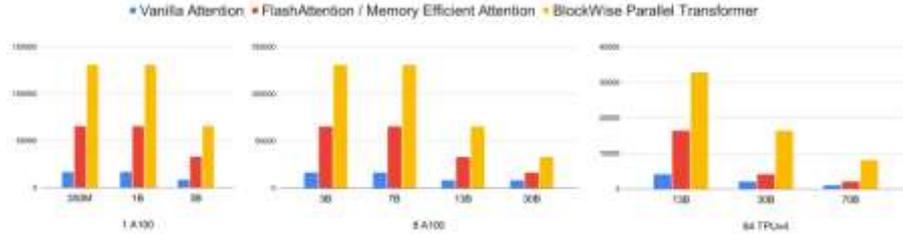

**Fig. 2.** Maximum context length during GPT training across three attention methods on different hardware setups (1×A100, 8×A100, 64×TPUv4). Blockwise Parallel Transformer (BPT) achieves up to 32× longer context windows than Vanilla Attention and 2–4× longer than Flash/Memory-Efficient Attention.

For Vietnamese, progress has been more gradual. PhoBERT [2] was an early success in monolingual pretraining, underscoring the importance of native-language resources. More recently, PhoRanker [3] showed that rerankers tailored for Vietnamese can outperform multilingual baselines. However, these models still face challenges in efficiency and robustness when deployed at scale.

## 3 Model and Experimental Setup

### 3.1 Experimental Setup and Resources

All experiments were conducted on a dedicated workstation running Ubuntu 20.04, equipped with a single CPU core, an NVIDIA A100 SXM4 GPU (80 GB memory), 256 GB RAM, and a 1 TB SSD. The model was implemented in PyTorch to leverage GPU acceleration and ensure reproducibility.

**Table 1.** Inference efficiency of ViRanker across different GPUs.

| Hardware (GPU) | Batch Size | Latency/Query (ms) | Peak Memory (GB) | Throughput (quesries/sec) |
|---|---|---|---|---|
| A100 80GB | 32 | 11.2 | 9.4 | 2,850 |
| V100 32GB | 16 | 21.7 | 11.1 | 1,320 |
| T4 16GB | 8 | 37.9 | 12.3 | 610 |

This setup provided sufficient memory for training with long sequences (up to 1024 tokens) and large batch sizes, while also enabling controlled efficiency tests across multiple GPUs. Table 1 reports inference latency and memory usage on A100, V100, and T4 devices, illustrating ViRanker's viability on both high-end and mid-tier hardware.

### 3.2 Data Preparation

To construct a large-scale Vietnamese reranking dataset, we assembled approximately 8 GB of raw text from Vietnamese Wikipedia, open-source GitHub repositories, and Vietnamese-language books and reports. Preprocessing included punctuation normalization (e.g., converting "hoà" → "hòa"), automated spelling correction with Gemma-1.5 and GPT-4o-mini, and sentence segmentation. Fragments under 512 tokens were discarded, while adjacent segments were merged to ensure document lengths up to 1024 tokens. This yielded around 3.5 million document chunks.

From this corpus, we generated query–document triplets using the Inverse Cloze Task: one sentence was sampled as the pseudo-query, while the remaining sentences formed the positive context. To increase difficulty, we applied a hybrid hard-negative mining strategy (Figure 2). For each query, BM25 retrieved the top 20 candidates, which were then reranked with BGE-M3 embeddings [6] and cosine similarity. To diversify distractors, we applied MMR, and selected the three most challenging non-relevant passages as negatives.

The final dataset contained 3.5 million triplets in the form:
$$\{query: str, pos: List[str], neg: List[str]\}.$$

Table 2 summarizes the dataset statistics, including document counts and average lengths across sources.

**Table 2.** Dataset statistics for ViRanker training corpus.

| Source | Size (GB) | # Documents | Avg. Tokens/Doc | Contribution |
|---|---|---|---|---|
| Vietnamese Wikipedia | 2.1 | 1.2M | 680 | General knowledge |
| Github (VN text repos) | 1.4 | 0.7M | 520 | Technical Text |
| Books and Reports | 4.5 | 1.6M | 980 | Long-form, formal |
| Total | 8.0 | 3.5M | ~730 | Balanced coverage |

### 3.3 Model Architecture

ViRanker builds on the BGE-M3 encoder, adapting it into a Vietnamese-specific cross-encoder reranker (Figure 1). Several key modifications were introduced to better handle Vietnamese text and improve deployment efficiency:

**Rotary Position Encoding (RoPE).** [7] Replaces absolute positional encoding, improving modeling of flexible word order and diacritics in Vietnamese. This enhances early-rank accuracy where fine-grained distinctions matter most.

**Blockwise Parallel Transformer (BPT).** [8] Substitutes FlashAttention [5] with a blockwise parallel mechanism that reduces memory usage while retaining accuracy, making ViRanker efficient on long documents and scalable across GPUs.

**Multi-Layer Perceptron (MLP) head.** Since BGE-M3 was originally designed as an embedding model rather than a reranker, we appended a lightweight scoring head to adapt outputs for query–document relevance.

**Triplet Ranking Loss.** [18] Training used triplet loss, aligning queries, positives, and hard negatives to enforce robust discriminative ranking. This loss has been widely used in retrieval and metric learning tasks to ensure effective separation of relevant and irrelevant candidates.

Together, these design choices allow ViRanker to balance linguistic sensitivity (RoPE), computational scalability (BPT), and robust learning (MLP + triplet loss), yielding a reranker tailored for Vietnamese information retrieval.

### 3.4 Hyperparameter Settings

ViRanker was trained with hyperparameters tuned for stability and efficiency (Table 3). We used a maximum sequence length of 1024 tokens, a large batch size of 512 for both training and evaluation, and 32 attention heads. Gradient checkpointing and accumulation were applied to reduce memory overhead, and a cosine learning-rate scheduler was used for smooth convergence.

This configuration balances large-scale contrastive training with practical GPU efficiency, enabling ViRanker to process long documents while avoiding excessive resource consumption.

Table 3. Training hyperparameters for ViRanker.

| Hyperparameter | Value | Hyperparameter | Value |
|---|---|---|---|
| # Attention Heads | 32 | Batch Size (train/eval) | 512/512 |
| Query Chunk Size | 32 | Max Sequence Length | 1024 |
| Key/Value Chunk Size | 32 | Learning Rate | $5 \times 10^{-5}$ |
| Memory Bank Size | 512 | Gradient Checkpointing | Enabled |
| Pooling Type | Mean | LR Scheduler | Cosine |
| # Training Epochs | 10 | Gradient Accumulation | 2 |

## 4 Experimental Results, Evaluation and Discussion

We evaluated ViRanker on the MMARCO-VI test set [4] using NDCG@k and MRR@k (k = 3, 5, 10), averaged across three random seeds. The results are summarized in Tables 4 and 5.

ViRanker achieves an NDCG@3 of 0.6815 and an MRR@3 of 0.6641, surpassing multilingual BGE rerankers [16, 17] and even outperforming PhoRanker [3] at the top of the ranking. These findings highlight its strength in early-rank retrieval—a critical factor for user satisfaction, as most users examine only the top few results [1]. Although PhoRanker retains a slight edge at NDCG@10, ViRanker remains highly competitive at deeper ranks. The extremely low scores of the fine-tuned PhoBERT reranker [2] (NDCG@3 = 0.0963) further emphasize the benefits of architectural innovations and careful data curation.

A closer analysis reveals how design choices contribute to these gains:
- **RoPE** [7] boosts early precision by capturing subtle word-order and diacritic cues unique to Vietnamese.
- **BPT** [8] improves long-document retrieval, sustaining performance at deeper cut-offs.
- **Hybrid hard-negative mining** strengthens robustness by training the model against near-duplicate distractors (BM25 [9] + MMR [10]).

**Table 4.** Comparison of NDCG scores on the MMARCO-VI Vietnamese test set. ViRanker achieves the strongest early-rank performance (NDCG@3 = 0.6815), outperforming all baselines including PhoRanker and BGE variants. Although PhoRanker remains slightly higher at NDCG@10, ViRanker maintains competitive deep-rank quality while substantially surpassing multilingual and PhoBERT-based rerankers.

| Model Name | NDCG@3 | NDCG@5 | NDCG@10 |
|---|---|---|---|
| namdp-ptit/ViRanker | **0.6815** | 0.6983 | 0.7302 |
| itdainb/PhoRanker | 0.6625 | **0.7147** | **0.7422** |
| kien-vu-uet/finetuned-phobert-passage-rerank-best-eval | 0.0963 | 0.1396 | 0.1681 |
| BAAI/bge-reranker-v2-m3 | 0.6087 | 0.6513 | 0.6872 |
| BAAI/bge-reranker-v2-gemma | 0.6088 | 0.6446 | 0.6785 |

**Table 5.** Comparison of MRR scores on the MMARCO-VI Vietnamese test set. ViRanker consistently outperforms baselines across all cut-offs, demonstrating reliable early-rank retrieval accuracy. The results confirm that architectural enhancements and the hard-negative mining strategy yield robust improvements in reciprocal-rank effectiveness.

| Model Name | MRR@3 | MRR@5 | MRR@10 |
|---|---|---|---|
| namdp-ptit/ViRanker | **0.6641** | **0.6894** | **0.7107** |
| itdainb/PhoRanker | 0.6458 | 0.6731 | 0.6830 |
| kien-vu-uet/finetuned-phobert-passage-rerank-best-eval | 0.0883 | 0.1131 | 0.1246 |
| BAAI/bge-reranker-v2-m3 | 0.5841 | 0.6062 | 0.6209 |
| BAAI/bge-reranker-v2-gemma | 0.5908 | 0.6108 | 0.6249 |

**Error Analysis.** Failures typically arise from (i) factoid-style queries (e.g., "Nguyễn Du năm sinh"), where ambiguity or entity overlap misleads the model, and (ii) long narrative documents, where diluted context reduces precision. Representative cases are shown in Table 6. These patterns suggest that while ViRanker excels on complex retrieval, it remains sensitive to brevity and long-context noise.

**Table 6.** Representative error cases for ViRanker on the MMARCO-VI dataset. Failures typically arise from ambiguous factoid-style queries or long narrative documents, highlighting areas for future robustness improvements.

| Query (translated) | ViRanker Top-1 Prediction | Correct Document | Likely Cause |
|---|---|---|---|
| "Nguyễn Du năm sinh" (Nguyen Du birth year) | A literary analysis article | Short factual query | Ambiguity in named entities |
| "Ảnh hưởng của mạng xã hội đến học sinh" (Impact of social media on students) | A news commentary | Survey-based research report | Long-document mismatch |
| "Kinh tế Việt Nam 2008" (Vietnamese economy 2008) | A blog opinion piece | Statistical report | Domain-specific terminology |

**Comparative Analysis with PhoRanker.** ViRanker outperforms PhoRanker [3] on descriptive, context-rich queries and medium-length documents (500–800 tokens). PhoRanker shows an edge on short, factoid queries. This indicates complementary strengths: PhoRanker is efficient for factual lookup, whereas ViRanker offers robustness across diverse query types.

**Limitations.** While ViRanker shows strong results, several limitations remain. The training corpus, though sizeable at 8 GB, is still modest compared to high-resource languages and may limit generalization to rare or specialized domains. Our analysis also revealed weaknesses on short factoid queries and very long documents, reflecting broader difficulties in long-context modeling [19]. In addition, the model's robustness to noisy inputs such as misspellings or missing diacritics has not been fully assessed. Although ViRanker runs efficiently on both high- and mid-tier GPUs, further optimizations like quantization or pruning will be necessary for deployment in resource-constrained settings.

Looking ahead, progress will likely depend on expanding Vietnamese training resources and adopting newer architectures that may surpass RoPE [7] and BPT [8] in handling long contexts. Parameter-efficient fine-tuning (PEFT) also offers a promising path for adapting ViRanker across domains without the heavy cost of retraining.

## 5   Conclusion

This paper introduced ViRanker, a Vietnamese-specific reranker that integrates the BGE-M3 backbone, Rotary Position Encoding (RoPE), and the Blockwise Parallel Transformer (BPT) [8]. Trained on a carefully curated 8 GB corpus with a hybrid negative sampling strategy, ViRanker achieves strong early-rank accuracy on the MMARCO-VI benchmark, outperforming multilingual baselines and remaining competitive with Vietnamese-specific systems such as PhoRanker.

Beyond accuracy, ViRanker is designed with efficiency in mind, running effectively on both high- and mid-tier GPUs, which makes it suitable for practical deployment. By releasing the model openly, we aim to foster reproducibility and encourage broader adoption in Vietnamese NLP research.

Looking forward, future work will focus on expanding the training corpus, exploring parameter-efficient fine-tuning, and investigating emerging architectures that may further improve efficiency and long-context handling. More broadly, our findings highlight the value of language-specific adaptation and rigorous preprocessing for advancing information retrieval in low-resource languages.